# ZeroSearch: Local Image Search from Text with Zero Shot Learning

Jatin Nainani
Department of Electronics and Telecommunication
K.J. Somaiya College of Engineering
Mumbai, India
jatin.nainani@somaiya.edu

Abhishek Mazumdar
Department of Electronics and Telecommunication
K.J. Somaiya College of Engineering
Mumbai, India
a.mazumdar@somaiya.edu

Viraj Sheth
Department of Electronics and Telecommunication
K.J. Somaiya College of Engineering
Mumbai, India
viraj.sheth@somaiya.edu

*Abstract*— The problem of organizing and finding images in a user's directory has become increasingly challenging due to the rapid growth in the number of images captured on personal devices. This paper presents a solution that utilizes zero shot learning to create image queries with only user provided text descriptions. The paper's primary contribution is the development of an algorithm that utilizes pre-trained models to extract features from images. The algorithm uses OWL to check for the presence of bounding boxes and sorts images based on cosine similarity scores. The algorithm's output is a list of images sorted in descending order of similarity, helping users to locate specific images more efficiently. The paper's experiments were conducted using a custom dataset to simulate a user's image directory and evaluated the accuracy, inference time, and size of the models. The results showed that the VGG model achieved the highest accuracy, while the Resnet50 and InceptionV3 models had the lowest inference time and size. The paper's proposed algorithm provides an effective and efficient solution for organizing and finding images in a user's local directory. The algorithm's performance and flexibility make it suitable for various applications, including personal image organization and search engines. Code and dataset for zero-search are available at: https://github.com/NainaniJatinZ/zero-search

## I. Introduction

Image search has become an essential tool for finding specific images from a large collection of images based on user-defined criteria. With the exponential growth of digital images, finding the right image has become increasingly difficult. Image search aims to solve this problem by providing efficient and accurate search results. Image search can be performed based on either text-based queries or image-based queries. Text-based image search allows users to search for images based on text keywords or phrases, while image-based search involves searching for similar images to a given image.

Both methods have their advantages and limitations, and choosing the right method depends on the user's specific needs. One of the main challenges in image search is the speed and accuracy of the search process. The time and space complexities of image search algorithms can make the search process slow and inefficient, making it difficult to search large collections of images. Additionally, image search algorithms must be able to accurately match images to user-defined criteria, which can be challenging due to variations in image content and background.

To address these challenges, we present an efficient and accurate image search algorithm that uses a combination of text and image-based queries. Our algorithm extracts features from the user's image directory using pre-trained deep learning models and uses a random image from the directory to identify relevant objects using object detection. If the identified object score is above a user-defined threshold, the image is used as a query for image-based search. If the identified object score is below the threshold, the algorithm repeats the process with a new random image until an appropriate query image is identified.

Our algorithm's main contribution is the ability to perform efficient and accurate image search using a combination of text and image-based queries. By leveraging pre-computed image feature arrays, we reduce the computational resources required for the search process while improving accuracy. Furthermore, by using object detection to identify relevant objects in the user's image directory, we can quickly and accurately identify query images, making the search process more efficient.

In conclusion, our algorithm provides an efficient and accurate solution to the challenges of image search, enabling users to quickly and easily find the right image from a large collection of images with the help of only text.

## II. Literature Review

Text-based image retrieval has become increasingly important in recent years, and researchers have proposed several innovative solutions to this problem. One study compared the performance of users when searching for information using photorealistic images and text, showing that image search was faster, less error-prone, and preferred by participants compared to text search. However, the study's limited scope and lack of discussion on potential biases call for caution when interpreting the results.

To facilitate research in this area, a Wikipedia-based Image Text (WIT) Dataset was introduced by Srinivasan et al [1], a large and diverse multimodal dataset with 37.5 million image-text pairs and 11.5 million unique images across 108 Wikipedia languages. Compared to other datasets, WIT has four main advantages: it is the largest multimodal dataset by the number of image-text examples, it is massively multilingual with coverage over 100+ languages, it represents a more diverse set of concepts and real-world entities, and it provides a challenging real-world test set. The dataset was created by extracting multiple different texts associated with an image from Wikipedia articles and Wikimedia image links, and rigorous filtering was applied to retain only high-quality image-text associations.

Another paper by Chawla et al [2], proposed a novel approach to Text Conditioned Image Retrieval (TCIR), where modified images are retrieved based on natural language feedback provided by users. The authors proposed to decompose an input image into its style and content features and apply modifications to the text feedback individually in both the style and content spaces before fusing them for retrieval. The results showed that this approach outperformed previous methods on the Fashion IQ dataset.

Additionally, a novel multi-granularity embedding learning (MGEL) model was proposed by Wang et al [3] for text-based person search, which aims to retrieve corresponding person images by a textual description from a large-scale image gallery. The proposed MGEL model generates multi-granularity embeddings of partial person bodies in a coarse-to-fine manner and performs multi-granularity image-text matching by integrating the partial embeddings at each granularity. The paper also introduces a part alignment loss to solve the ambiguity embedding problem and demonstrate the effectiveness of the proposed method through extensive experiments on a public benchmark.

Finally, a new approach for language-based search of large-scale image and video datasets using transformers was proposed by Miech et al [4]. The approach consists of mapping text and vision to a joint embedding space, also known as dual encoders. However, this approach can be inefficient for large-scale datasets, and the authors proposed a solution that incorporates fast nearest neighbor search with inverted indexing to speed up the search process.

Overall, these studies highlight the importance of text-based image retrieval and propose innovative solutions to this problem. The WIT dataset, in particular, provides a valuable resource for researchers in this area, and the proposed models show promising results in improving retrieval performance and providing a more accurate and efficient user experience.

### III. PROPOSED METHODOLOGY

*A. Block Diagram*

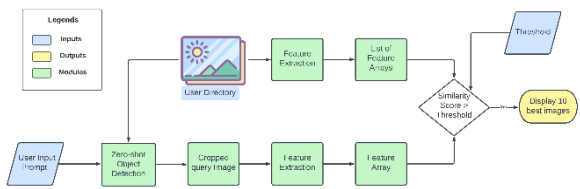

*Fig 1: Flow Chart*

The above diagram demonstrates the methodology followed by the solution. It will be further discussed in section C.

*B. Architecture*

   *1) Zero shot object detection*

Zero shot object detection is a method that enables object detection for classes that are not present in the training set. It leverages semantic information to classify novel classes. Text-conditioning is a technique that helps to generate a better embedding of the object class, thus improving the accuracy of zero shot object detection. It involves conditioning the model with natural language descriptions of the object classes. This allows the model to learn the semantic relationships between different classes, which can help to recognize and classify unseen classes. Overall, zero shot object detection with text-conditioning is a promising approach for detecting and classifying a wide range of objects, even if they have not been seen during training.

   *a) OwlViT - Vision Transformer for Open-World Localization*

OwlVIT [5] combines the Transformer-based Vision Transformer (ViT) architecture with natural language processing (NLP) techniques to enable object detection based on natural language descriptions without the need for any training data.

The architecture of OwlVIT, shown in Fig. 2, consists of a ViT model that takes in the image and an NLP model that processes the textual input. The image features are extracted by the ViT model, and the textual input is processed by the NLP model. The two streams are then concatenated and fed through a few fully connected layers to predict the bounding boxes and class labels. The ViT model used in OwlVIT is pre-trained on large-scale image datasets, such as ImageNet, to learn rich image representations.

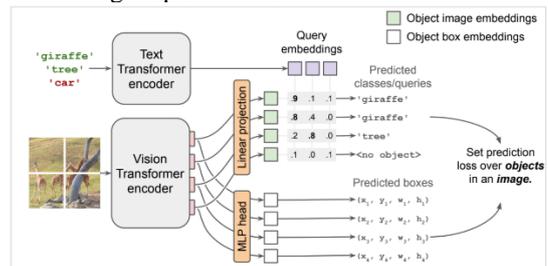

*Fig 2: Architecture for Zero Shot Object Detection*

In terms of training, OwlVIT is trained using a combination of image and text data. The image data is used to pre-train the ViT model, while the text data is used to fine-tune the model for zero-shot object detection. The text data consists of natural language descriptions of objects, which are used to condition the model for object detection. During inference, the model can detect objects based on natural language descriptions that it has not seen during training.

As for the evaluation metrics, the authors of OwlVIT used standard object detection metrics such as precision, recall, and F1 score to evaluate the model's performance. They also compared the results of OwlVIT with other state-of-the-art object detection models, such as Faster R-CNN and RetinaNet, on benchmark datasets such as COCO and VG. The results showed that OwlVIT outperformed these models in terms of zero-shot object detection accuracy.

In summary, OwlVIT is a novel zero-shot text-conditioned object detection model that combines the ViT [6] architecture with NLP techniques to enable object detection based on natural language descriptions. The model is pre-trained on large-scale image datasets and fine-tuned on text data for zero-shot object detection. The model's performance is evaluated using standard object detection metrics and compared with other state-of-the-art models on benchmark datasets.

   *2) Feature Extraction*

Image feature extraction is a fundamental technique in computer vision that involves extracting useful information from an image that can be used to perform various tasks, such as object recognition, image retrieval, and image classification. The goal of feature extraction is to transform an image into a set of features that are more meaningful and easier to analyze than the original image. Feature extraction is often done using pre-trained models, which are deep neural networks trained on large image datasets such as ImageNet.

The process of image feature extraction using pre-trained models involves passing an image through a deep neural network, such as VGG, ResNet, or Inception, and extracting the activations from one of the layers in the network. These activations, also called feature maps, represent different visual patterns, such as edges, corners, and textures, that are present in the image. The feature maps can be used to create a feature vector, which is a numerical representation of the image that captures its visual content. This feature vector can then be used for various tasks, such as image classification, object detection, and image retrieval.

One of the advantages of using pre-trained models for feature extraction is that they have already learned to extract high-level features from images. These models are trained on large datasets and have already learned to recognize common visual patterns that are useful for various tasks. By using pre-trained models for feature extraction, researchers can save time and computational resources by not having to train their own models from scratch. Additionally, pre-trained models can be fine-tuned on smaller datasets to adapt to specific tasks, which can further improve their performance.

In conclusion, image feature extraction using pre-trained models is a powerful technique in computer vision that involves extracting useful information from an image that can be used for various tasks. The process involves passing an image through a deep neural network and extracting the activations from one of the layers in the network. The advantages of using pre-trained models for feature extraction include their ability to extract high-level features, their pre-trained weights that save time and computational resources, and their adaptability to specific tasks through fine-tuning.

*3) Object Detection Models*

In this work, the feature extraction process is carried out by below mentioned pretrained models.

*a)* VGG16 [7] is a convolutional neural network that consists of 16 layers. It is a popular model for image recognition tasks and was trained on the ImageNet dataset. The input size for VGG16 is 224 x 224 pixels, and it has 138 million parameters. VGG16 is known for its high accuracy on image classification tasks, but it is computationally expensive and has a large memory footprint.

*b)* ResNet50 [8] is a deep neural network that consists of 50 layers. It was designed to address the problem of vanishing gradients in deep neural networks by using skip connections. ResNet50 was trained on the ImageNet dataset, and its input size is also 224 x 224 pixels. ResNet50 has 25 million parameters, making it relatively lightweight compared to VGG19. ResNet50 is known for its high accuracy and efficiency on image classification tasks.

*c)* MobileNetV2 [9] is a neural network that is designed to be highly efficient and lightweight. It was trained on the ImageNet dataset, and its input size is also 224 x 224 pixels. MobileNetV2 has 3.4 million parameters and is known for its speed and efficiency on image classification tasks. MobileNetV2 uses depthwise separable convolutions to reduce the number of parameters and improve computational efficiency.

*d)* InceptionV3 [10] is a deep neural network that was designed to address the problem of vanishing gradients in deep neural networks. It was trained on the ImageNet dataset, and its input size is also 224 x 224 pixels. Inception has 27 million parameters and is known for its high accuracy on image classification tasks. Inception uses a combination of convolutional layers and pooling layers to extract features from images. It also uses a novel module called the "Inception module" that allows the network to learn features at multiple scales.

*4) Similarity Score*

The cosine similarity matrix is a common way to compare feature arrays extracted from images using pre-trained models. It is a mathematical technique that measures the similarity between two feature vectors by computing the cosine of the angle between them. The cosine similarity ranges from -1 to 1, where a value of 1 indicates that the two vectors are identical, 0 indicates that they are completely dissimilar, and -1 indicates that they are opposite to each other.

$$CosineSimilarity(array\ A, array\ B) = \frac{A \cdot B}{\|A\| \times \|B\|}$$

*Equation 1: Cosine Similarity Score*

In the context of image feature extraction, the cosine similarity matrix is often used to measure the similarity between two images based on their feature vectors. The feature vectors are first extracted using a pre-trained model, such as VGG, ResNet, or Inception, and then compared using the cosine similarity matrix. The resulting matrix is a square matrix where each element represents the cosine similarity between two images.

The cosine similarity matrix is useful for various tasks, such as image retrieval and image clustering. For image retrieval, the cosine similarity matrix can be used to find the most similar images to a given query image by ranking the images based on their cosine similarity scores. For image clustering, the cosine similarity matrix can be used to group similar images together based on their cosine similarity scores.

One of the advantages of using the cosine similarity matrix is that it is computationally efficient and can handle high-dimensional feature vectors. Additionally, the cosine similarity matrix is invariant to scaling, which means that it is not affected by the magnitude of the feature vectors. This makes it a robust technique for comparing feature vectors that may have different magnitudes.

## C. Algorithm

### 1) Stepwise Implementation of Main Process

**Algorithm 1**: Local Image Search with Text

**Inputs**: User text input (prompt), User directory (file_path), Threshold (thresh)
**Output**: List of Image paths (output_images)

```
START
    feature_array ← extract_features(file_path)    // extract features of all images in directory
    image ← random(file_path)                      // get a random image
    bounding_box, confidence ← object_detection(image, prompt, threshold)
    WHILE length(bounding_box) == 0 DO             // retry till prompt detected in image
        image ← random(file_path)
        bounding_box, confidence ← object_detection(image, prompt, threshold)
    query_image ← image[bounding_box]              // crop the image
    query_feature ← extract_features(query_image)
    similarity_scores ← cosine_similarity(query_feature, feature_array)
    output_images ← get_top_10_images(similarity_scores)
                                                   // retrive path of images with highest scores
STOP
```

Algorithm 1 is designed to search for images within a user-specified directory (folder for PC or gallery for phone) based on a user's text input. The default folder for the search is Pictures. The algorithm uses a cosine similarity metric to sort the list of images in descending order. The input for the algorithm includes the user directory, default folder, user text input (e.g., shoe, phone, etc.), and a threshold value for the input. The output of the algorithm is a list of images sorted in descending order of cosine similarity. The list contains the top 10 best results.

The implementation of the algorithm can be done using Python as the primary language. The algorithm can be integrated with a user-friendly GUI built with the help of the Tkinter library. The GUI can provide an interface for the user to select the directory and input text, and also display the results of the search.

In order to extract features from images, the algorithm can use different pre-trained models from the Keras applications library, such as ResNet50 or InceptionV3. These models can be fine-tuned to extract features specific to the user's image directory, thus improving the accuracy of the search.

**Algorithm 2**: Zero Shot Object Detection

**Inputs**: Image (img), User text input (prompt), Threshold (thresh)
**Output**: Bounding box (bbox), confidence score (conf)

```
START
    results ← model(prompt, img)                   // inference model
    conf ← -100                                    // set initial confidence
    FOR box, score in results DO                   // iterate through the results
        IF (score > thresh) AND (score>conf) THEN
                                                   // score needs to be greater than threshold
                                                   // and better than previous
            bbox ← box
            conf ← score
    RETURN bbox, conf                              // return the best bouding box and its
                                                   // confidence
STOP
```

To perform object detection on the selected image, the transformers library can be used to implement zero-shot object detection. Algorithm 2 shows this process. This approach allows the algorithm to detect objects in an image without requiring a specific object detection model. By using the transformers library, the algorithm can detect objects in the image and extract the relevant features to compare with other images in the directory.

Once the query image is selected and its features are extracted, the algorithm can use a simple linear for loop to compare the features of each image in the directory with the query image. The similarity between images can be calculated using cosine similarity, which measures the angle between two vectors. The list of images can then be sorted in descending order of similarity and the top 10 best results can be displayed to the user.

Overall, by using Python, Tkinter, Keras applications, and the transformers library, the implementation of the algorithm can provide a user-friendly interface and accurate results for image search within a specified directory.

### 2) Efficiency

Using image feature arrays is more efficient than just inferencing the large OWL model on every image in the user directory for every single prompt due to several reasons.

Firstly, the OWL model is a complex deep learning model that requires significant computational resources to run, especially when dealing with a large number of images. Running the OWL model on every image in the user directory for every single prompt would be computationally expensive and time-consuming, making the image search process slow and inefficient.

On the other hand, image feature arrays are pre-computed arrays of numerical features extracted from images using deep learning models. These features can be easily computed and stored offline, and can be used for efficient image search without requiring the OWL model to be run on every image in the directory.

When a user enters a search query, the algorithm can use the pre-computed image feature arrays to quickly find the most similar images in the directory without needing to run the OWL model on every image. This approach significantly reduces the computational resources required for the search process, making it more efficient and faster.

Moreover, the use of pre-computed image feature arrays can also improve the accuracy of the search results. Deep learning models used to compute these features are often trained on large datasets, making them robust to variations in image content and background. As a result, using pre-computed image feature arrays can lead to more accurate image search results, especially when dealing with large image datasets.

Thus, using pre-computed image feature arrays is more efficient and accurate than just inferencing the large OWL model on every image in the user directory for every single prompt. This approach allows for faster image search while also requiring fewer computational resources.

### 3) Need for Randomness

Passing a random image through OWL was done to ensure that the algorithm selects an image that contains the relevant object or objects mentioned in the user's input text.

The initial objective of passing a random image through OWL was to check if the object detection tool can identify the object or objects in the image with a score that meets the threshold value set by the algorithm. If the score is above the threshold, the algorithm proceeds with the selected image as the query image.

However, if the score is below the threshold, the algorithm selects another random image and repeats the process. This iterative process ensures that the algorithm selects an image that contains the relevant object or objects mentioned in the user's input text with a high score, rather than selecting an image that does not contain the relevant object or objects.

Thus, passing a random image through OWL helps to ensure that the algorithm selects an image that contains the relevant object or objects mentioned in the user's input text, improving the accuracy of the search results.

## IV. EXPERIMENTATION

In this section, the experimentation process undertaken to validate the model shall be discussed along with a detailed outlook on the model performance and results achieved. The dataset collected to feed the models and evaluation metrics utilized to measure model performance shall also be touched upon.

### A. Dataset

In this work, we have manually extracted a diverse set of 152 images across the web. Since the OWL-ViT is modelled over generalized datasets, the manual images collected included objects across -

1) Animals (cats, dogs, etc)
2) People (man, woman, family, etc)
3) Streets (roads, cars, etc)
4) Food
5) Statues
6) and generalised images such as monuments, balloons, etc.

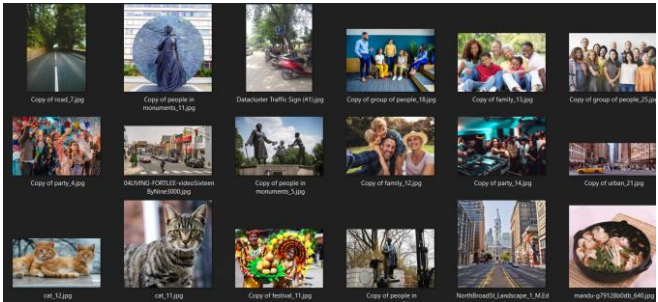

*Fig 3: Small subset of dataset used*

This enabled a diverse set of image types on which the model can be implemented and the efficacy of the model can be evaluated on diverse sets of results. Each of these images not only catered to particular category of prompts but also contained varied objects to search for disparate classes of objects. After creating our dataset, we validated it to make sure the information was accurate and ready for analysis.

### B. Evaluation Metrics

In this sub-section, the evaluation metrics used to quantize the model performance shall be discussed. This work is based on a 'Classification Problem'. Hence, we have used Accuracy as an metric to evaluate the model performance.

Accuracy is a common evaluation metric used to measure the performance of a classifier in a supervised learning task. It measures the proportion of correctly classified instances over the total number of instances in the dataset.

$$Accuracy = \frac{Number\ of\ correct\ predictions}{Total\ number\ of\ predictions}$$

*Equation 2: Accuracy for zero-search*

Classification accuracy is a useful metric for comparing different classifiers or evaluating the performance of a single classifier. However, it may not be appropriate in all cases, especially when the dataset is imbalanced or when the cost of misclassifying certain instances is high.

Another important metric used in this work revolved around understanding the time complexity of the solution. Time per inference step for transfer learning models refers to the amount of time it takes for the model to make a prediction or generate output on a new input during inference, after the model has been fine-tuned on a specific task or dataset.

In transfer learning, a pre-trained model is fine-tuned on a new task or dataset by training the model with additional data specific to the new task. After the model has been fine-tuned, it can be used for inference on new inputs.

The time per inference step refers to the amount of time it takes for the fine-tuned model to process a single input and generate output. This time is important for assessing the performance and efficiency of the model, and it can be influenced by factors such as the model architecture, the size of the fine-tuned model, the hardware being used for inference, and the complexity of the task or input.

Finally, to tackle space complexity of the model, the sizes of the pretrained models used was also done.

### C. Experimentation Methodology

In this sub-section, the proposed experimentation process which was implemented to test the model efficiency has been detailed.

The testing process encompassed a collection of multiple text inputs, some of them being - Cat, Dog, Balloons, Food, Road, Statue, Man, Family, Woman, Car & Glasses; and threshold inputs and comparing the performance of multiple transfer learning models - VGG16, InceptionV3, Xception, Resnet50 & MobileNetV2; based on the evaluation metrics defined above. As the methodology process flow involves choosing a random image as the query image and extracting features of the query image and comparing it with the feature extracts of each of the images in the fed folder to obtain the similarity score, it can be understood that accuracy of the model somewhat depends on the query image chosen. Since the selection of the query image is random, to enhance the robustness of the solution we have averaged the results for a prompt for multiple thresholds. This action eliminates the bias that a certain image might provide better features and hence providing better results irrespective of the model used to extract the feature providing a realistic reflection of the model performance.

The above testing was performed on 2 system specifications. First, with only a CPU - Intel(R) Core (TM) i5-1035G1 CPU @ 1.00GHz 1.19 GHz. This was done to test the solution's performance on low end systems. Second, with a GPU - 1xTesla K80, compute 3.7, having 2496 CUDA cores, 12GB GDDR5 VRAM.

In summary, with the usage of multiple search prompts and thresholds the efficiency to extract appropriate image features have been measured on the basis of the evaluation metrics defined. Lastly, time and space complexity of the approach was also noted for each model used.

## V. RESULTS AND ANALYSIS

Table 1 presents a comparison of the accuracy based on various user prompts. Several prompts were tested, some of their results are shown in Table 1. From the results, it is evident that the accuracy of the models varies based on the prompt and the type of model used. For instance, for prompts such as 'food' and 'road,' the VGG model has the highest accuracy of 1 and 0.94, respectively, while the MobileNetV2 model has the lowest accuracy of 0.72 and 0.52, respectively. On the other hand, for prompts such as 'dog' and 'family,' the Resnet50 model has the highest accuracy of 0.9 and 0.87, respectively.

*Table 1: Prompt Results*

| Prompt | Model | Accuracy |
|---|---|---|
| Cat | VGG | 0.91 |
|  | Resnet50 | 0.84 |
|  | MobileNetV2 | 0.61 |
|  | InceptionV3 | 0.85 |
| Balloons | VGG | 0.8 |
|  | Resnet50 | 0.95 |
|  | MobileNetV2 | 0.75 |
|  | InceptionV3 | 0.8 |
| Food | VGG | 1 |
|  | Resnet50 | 0.9 |
|  | MobileNetV2 | 0.72 |
|  | InceptionV3 | 0.96 |
| Road | VGG | 0.94 |
|  | Resnet50 | 1 |
|  | MobileNetV2 | 0.52 |
|  | InceptionV3 | 0.95 |
| Statue | VGG | 0.81 |
|  | Resnet50 | 0.75 |
|  | MobileNetV2 | 0.87 |
|  | InceptionV3 | 0.62 |
| Family | VGG | 0.91 |
|  | Resnet50 | 0.87 |
|  | MobileNetV2 | 0.86 |
|  | InceptionV3 | 0.74 |

Furthermore, Table 2 displays the averaged results from Table 1. It even supports the results with Metrics like Time for Inference and Size of the model, which vary significantly based on the type of model used. The VGG model has the highest time of inference and size, with a value of 69.5 ms and 528 MB, respectively, for all prompts. In contrast, the Resnet50 and InceptionV3 models have the lowest time of inference and size, with values of 58.2 ms and 98 MB, respectively, for all prompts.

Five outputs for our GUI application are also shown in Fig 4.

In conclusion, the results of the study demonstrate that the choice of transfer learning model can significantly impact the accuracy, time of inference, and size of the model. Therefore, researchers should consider the trade-off between accuracy and model size when selecting a model for a particular application. Additionally, the results suggest that Resnet50 and InceptionV3 models can be used as alternatives to VGG, considering their lower time of inference and size without significant reductions in accuracy.

*Table 2: Comparative Analysis of Models*

| Model | Avg. Accuracy | Time of Inference (CPU) | Time of Inference (GPU) | Size (MB) |
|---|---|---|---|---|
| VGG19 | 0.895 | 69.5 | 4.2 | 528 |
| Resnet50 | 0.885 | 58.2 | 4.6 | 98 |
| MobileNetV2 | 0.722 | 25.9 | 3.8 | 14 |
| InceptionV3 | 0.82 | 42.2 | 6.9 | 92 |

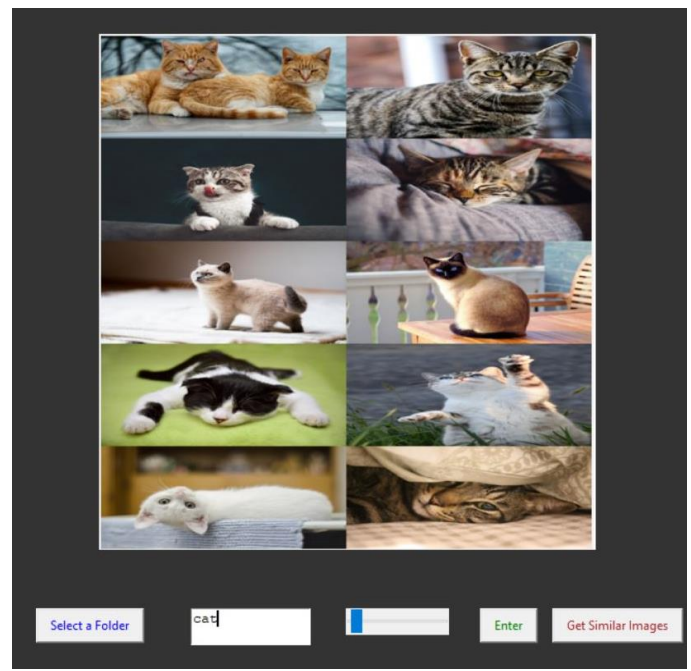

*(a)*

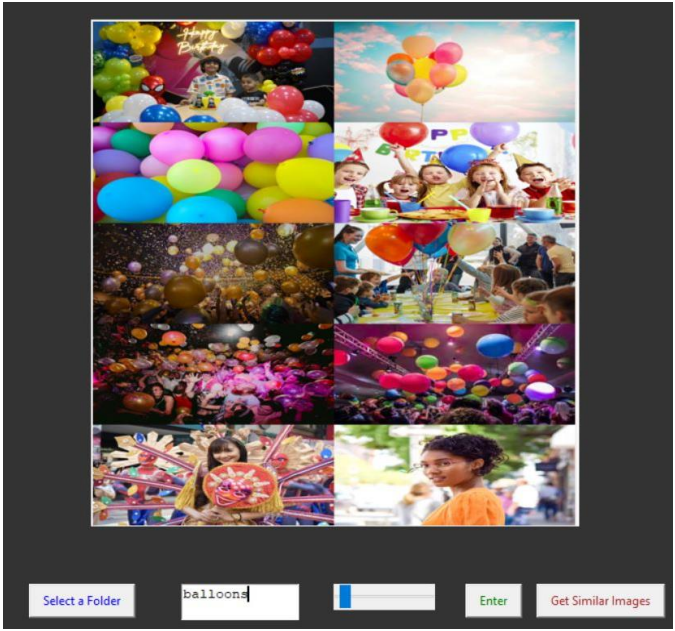
*(b)*

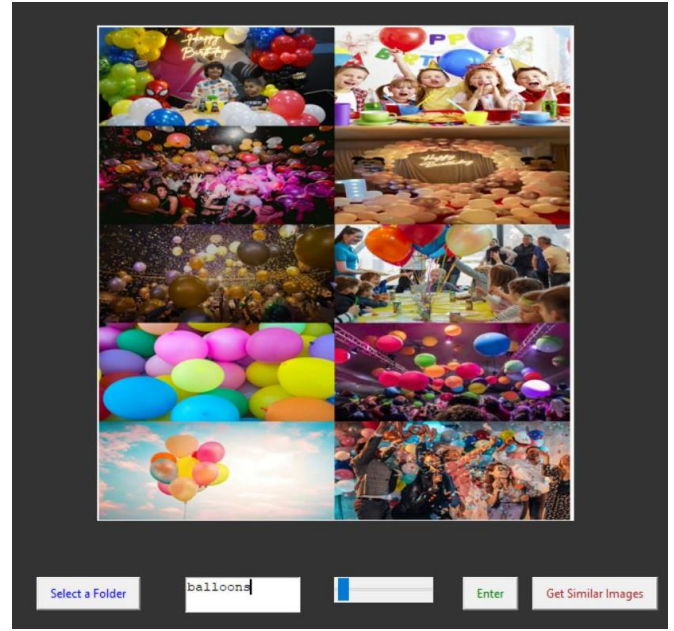
*(c)*

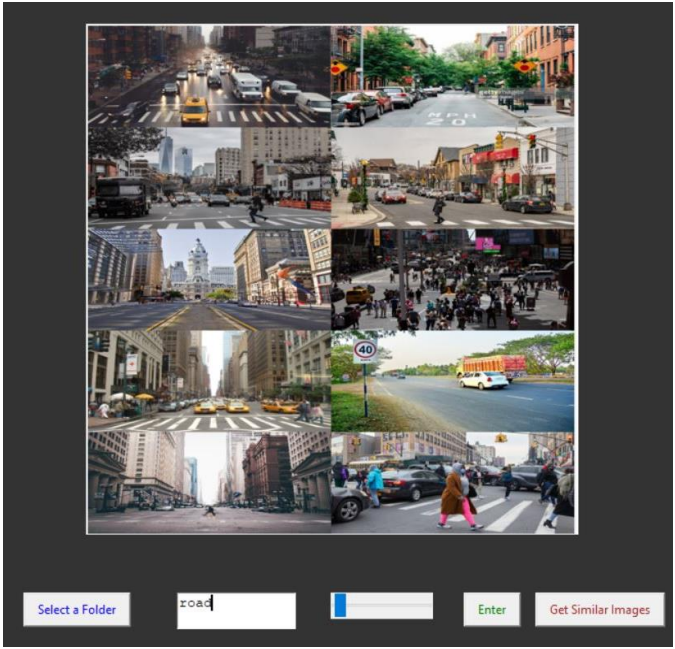
*(d)*

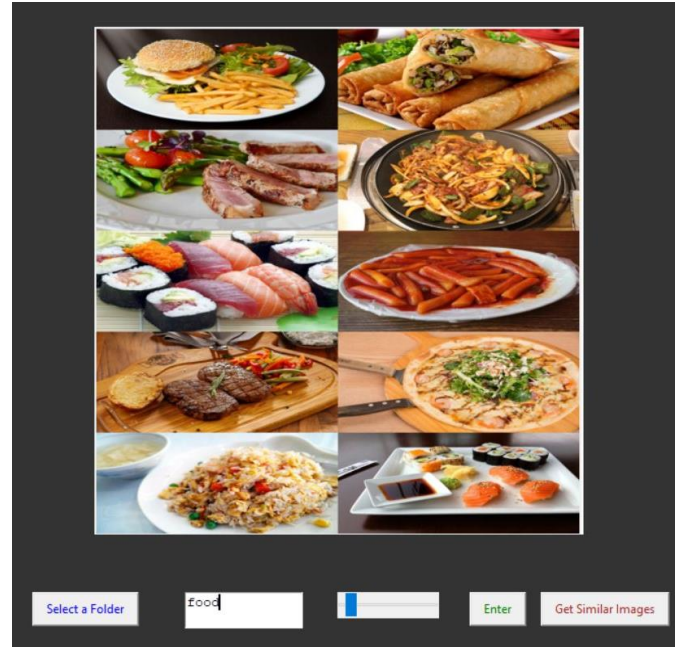
*(e)*

*Fig 4: (a) VGG16 Output for "CAT" (b) Resnet50 Output for "BALLOONS" (c) MobileNetV2 Output for "BALLOONS" (d) InceptionV3 Output for "ROAD" (e) VGG16 Output for "FOOD"*

## VI. Conclusion

In this paper, we have presented an algorithm that can efficiently retrieve images from a user directory based on a user's input query. Our algorithm uses transfer learning models to extract features from images and calculates the cosine similarity between them to sort the images in descending order of relevance.

The problem of image retrieval is becoming increasingly important as people are generating vast amounts of image data every day. With the help of our algorithm, users can quickly find relevant images from their personal photo galleries or folders, making it a valuable tool for personal and professional use.

Our contribution to this problem is two-fold. First, we have presented a novel approach to retrieve images based on user input that utilizes transfer learning models for efficient feature extraction. Second, we have demonstrated the trade-off between accuracy and model size and suggested that researchers should carefully consider the selection of transfer learning models for different applications based on their specific needs.

Overall, our algorithm shows promising results in terms of accuracy, speed, and size. In conclusion, we believe that our approach can help researchers and developers to create more efficient image retrieval systems and contribute to the advancement of this field.

## VII. FUTURE WORK

While pre-trained models are designed to work well on a variety of tasks, fine-tuning them to your specific use case can improve their accuracy and speed. You can experiment with different architectures, such as ResNet, Inception, or EfficientNet, and adjust their hyperparameters, such as learning rate, batch size, or weight decay. You can also try different training techniques, such as transfer learning, self-supervised learning, or adversarial training, to improve the performance of your model.

User feedback can help you improve the relevance of your search results over time. You can ask users to rate the relevance of the retrieved images on a scale from 1 to 5, for example, and use that feedback to update and improve your search algorithm. You can also use active learning techniques to selectively query users for feedback on specific images that the algorithm is uncertain about.

In addition to deep learning features, other types of visual features can improve the accuracy of your app, such as color histograms, texture features, or edge detection. You can also combine different types of features to achieve better results. For example, you can use a combination of deep learning features and color histograms to capture both semantic and visual information about the images.

Finally, consider integrating your app with other tools and services: Image search is often just one part of a larger workflow or use case. By integrating your app with other tools and services, such as image tagging, object recognition, or content management systems, you can provide a more complete solution for your users. For example, you can allow users to tag images with custom labels or descriptions, or automatically recognize objects in the images and provide additional information or actions based on those objects.

## ACKNOWLEDGMENT

We would like to express my very great appreciation to Dr. Rupali Patil for their valuable and constructive suggestions during the planning and development of this project. Their willingness to give their time so generously has been very much appreciated. We would also like to extend our gratitude to the entire team of HuggingFace team as well as that of Keras for implementing and maintaining their framework, without which this project would not be possible.